# Search-based Methods to Bound Diagnostic Probabilities in Very Large Belief Nets

**Max Henrion**

Rockwell International Science Center
Palo Alto Laboratory
444 High Street, #400
Palo Alto, Ca 94301
Henrion@Sumex-AIM.Stanford.EDU

## Abstract

Since exact probabilistic inference is intractable in general for large multiply connected belief nets, approximate methods are required. A promising approach is to use heuristic search among hypotheses (instantiations of the network) to find the most probable ones, as in the TopN algorithm. Search is based on the relative probabilities of hypotheses which are efficient to compute. Given upper and lower bounds on the relative probability of partial hypotheses, it is possible to obtain bounds on the absolute probabilities of hypotheses. Best-first search aimed at reducing the maximum error progressively narrows the bounds as more hypotheses are examined. Here, qualitative probabilistic analysis is employed to obtain bounds on the relative probability of partial hypotheses for the BN2O class of networks networks and a generalization replacing the noisy OR assumption by negative synergy. The approach is illustrated by application to a very large belief network, QMR-BN, which is a reformulation of the Internist-1 system for diagnosis in internal medicine.

## 1 INTRODUCTION

Bayesian belief networks provide a tractable basis for expressing uncertain knowledge at both qualitative and quantitative levels, in a way that is formally sound and intuitively appealing. They are already being used in a wide variety of applications, including knowledge bases of up to about one thousand nodes. A major obstacle to their application for still larger applications is the limitations of available algorithms for diagnostic inference. Exact diagnostic inference in general belief networks has been shown to be NP-hard (Cooper, 1991). Hence, there is considerable interest in the development of methods that provide greater efficiency at the cost of imprecision in the results (Henrion, 1990b).

There have been two main directions in which researchers have sought efficient approximate algorithms. One approach involves random sampling of network instantiations, also known as stochastic simulation (Henrion, 1988). The other involves search among the space of instantiations (hypotheses) to find those that are most probable. Cooper (1984) employed this approach in Nestor, to obtain the most probable hypotheses. Peng and Reggia (1987a, 1987b) and Henrion (1990a) developed more powerful admissability heuristics to prune the search tree, allowing more efficient search of BN2O networks, that is bipartite networks consisting of independent diseases, conditionally independent findings, and noisy ORs, as described in Section 3. These methods are guaranteed to find the most probable composite hypotheses, and their relative probabilities (ratio of posterior probabilities of hypotheses). Peng and Reggia (1989) and Henrion (1990a) also describe methods to bound the absolute probabilities of the composite hypotheses.

Peng and Reggia's approach to abductive reasoning is based on the notion of minimal covering sets of diseases which explain observed findings. They use logical techniques initially to identify covering sets for the given findings, and then use probabilistic methods to find the most probable hypotheses. This scheme assumes *zero leaks*, that is that no findings can occur "spontaneously" in the absence of any explicitly modelled cause. For the QMR-BN application to be described here, and indeed most medical problems, most findings have non-zero leak rates due to false positives, and so an adequate diagnosis does not necessarily all have to explain all observed findings. This makes the covering set approach inapplicable.

Shimony and Charniak (1990) describe a search-based method that finds the MAP (Maximum A-posteriori Probability) assignments to general belief networks. They show how any belief network can be converted to an equivalent weighted boolean function DAG, and that solving the best selection problem (minimum cost assignment) for this network is equivalent to finding the MAP assignment for the belief network. While the best selection problem is also NP-hard, standard best-first search can be relatively efficient in practice.

If the results of diagnostic inference or abductive reasoning are to be used as the basis for making decisions, for example how to treat a patient, or what additional tests to order, knowing the relative probabilities of the most likely complete hypothesis is not enough. We want to know the absolute probabilities, or at least have bounds on them, and we want often want to know the marginal posterior probabilities of individual diseases, or of one or two diseases, rather than of complete assignments which include instantiations of all the other nodes.

To obtain bounds on the absolute probabilities, we need bounds on the relative probabilities of all the hypothesis that we have not explicitly examined in the search. That is we want to find bounds on the sum of the relative probabilities of the possible extensions of a given hypothesis. Given bounds on the relative probabilities of all hypotheses, we can compute bounds on the absolute probabilities. However, to find such bounds requires additional knowledge of properties of the network. Qualitative knowledge about influences (Wellman, 1990; Wellman & Henrion, 1991) is a useful source of information to obtain bounds, as we shall see.

This paper presents improvements and generalizations to the TopN algorithm. First, I will describe the QMR-BN belief network which is the application providing a context and motivation for this work on algorithm design. I then describe a generalization of the noisy-OR assumption of the BN2O networks, to negative product synergy. This forms a basis for generalized bounding theorems, including a new lower bound, that provides a significant improvement on TopN as presented in Henrion (1990b). Qualitative probabilistic analysis, using signs of influence and synergies, provides a clearer and more general basis for obtaining these. I then describe a method to obtain bounds on the posterior probability of hypotheses and for individual diseases. Finally, I present results from application to the QMR-BN network, showing progressive improvement as search is extended.

## 2 QMR AND INTERNIST-1

QMR (Quick Medical Reference) is a knowledge-based system for supporting diagnosis by physicians in internal medicine (Miller et al, 1986). It is a successor to the Internist-1 system (Miller et al, 1982). The version of the knowledge-base used here contains information for 576 diseases (of the estimated 750 diseases comprising internal medicine) and over 4000 manifestations, such as patient characteristics, medical history, symptoms, signs, and laboratory results. In this paper, these are referred to generically as *findings*. QMR contains over 40,000 disease-finding associations. It represents about 25 person-years of effort in knowledge engineering and is one of the most comprehensive structured medical knowledge-bases currently existing.

The knowledge-base consists of a *profile* for each disease, that is, a list of the findings associated with it. Each such association between disease $d$ and finding $f$ is quantified by two numbers: The *evoking strength* is a number between 0 and 5 which answers the question "Given a patient with finding $f$, how strongly should I consider disease $d$ to be its explanation?". The *frequency* is a number between 1 and 5 answering the question "How often does a patient with disease $d$ have finding $f$?". Associated with each finding $f$ is an *import*, being a number between 1 and 5 answering "To what degree is one compelled to explain the presence of finding $f$ in any patient?".

## 3 QMR-BN: A PROBABILISTIC INTERPRETATION OF QMR

The aim of this project[1] is to develop a coherent probabilistic interpretation of QMR, which we call QMR-BN (for Belief Network), and eventually a version with treatment decisions and cost or value models, which we call QMR-DT (for Decision Theory). The first goal is to improve the consistency of the knowledge base and to explicate the independence assumptions it incorporates. A second goal is to provide a challenging example to develop and test new algorithms for probabilistic reasoning. The current version is a reformulation of the Internist-1 knowledge-base. See Henrion (1990a), Shwe et al, (1991) and Middleton et al, (1991) for more details.

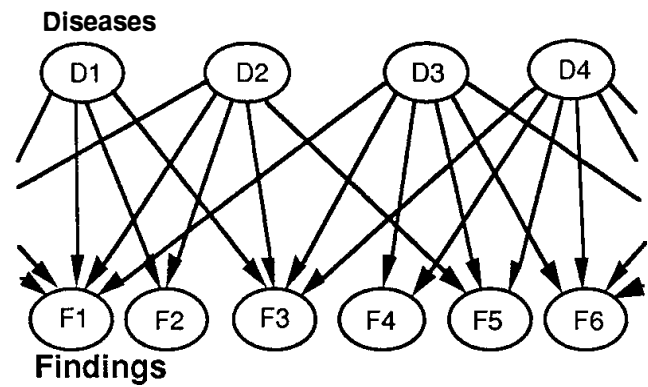

Figure 2: BN2O Belief net

A probabilistic representation can be divided into two aspects: The framework of qualitative assumptions about dependence and independences, and the quantification of the probabilities within that framework. QMR-BN currently follows INTERNIST-1 and QMR in assuming that all diseases and findings are binary variables, being either present or absent, without intermediate values. The initial qualitative formulation incorporates the following assumptions, expressed by the belief net in Figure 2:

**Assumption 1 (MID):** Diseases are marginally independent.

**Assumption 2 (CIF):** All findings are conditionally independent of each other given any hypothesis.

---

[1] This project is a collaboration with Gregory Cooper, David Heckerman, Eric Horvitz, Blackford Middleton, and Michael Shwe.

**Assumption 3 (LNOG):** The effects of multiple diseases on a common finding are combined as a *Leaky Noisy OR Gate*. Suppose $S_{df}$ is the link event that disease $d$ is sufficient to cause finding $f$.[2] The noisy OR assumption is that finding $f$ will occur if any link event occurs linking a present disease to $f$, and that these link events are independent. (This is sometimes known as *causal independence*.) With a *leaky* noisy OR an additional leak event $L_f$ is possible, which can cause $f$ to occur even with no explicit disease present.

**Definition 1 (BN2O):** The class of bipartite belief nets conforming to Assumptions 1, 2 and 3, are termed BN2O.

Some of the findings in INTERNIST-1, such as the demographics or family history of a patient, are not actually caused by diseases, but rather circumstances or risk factors that may affect disease probabilities. These variables should rearranged for ease of assessment so that they influence the diseases rather than *vice versa*. Currently, we have done this with age and sex as represented in figure 3.

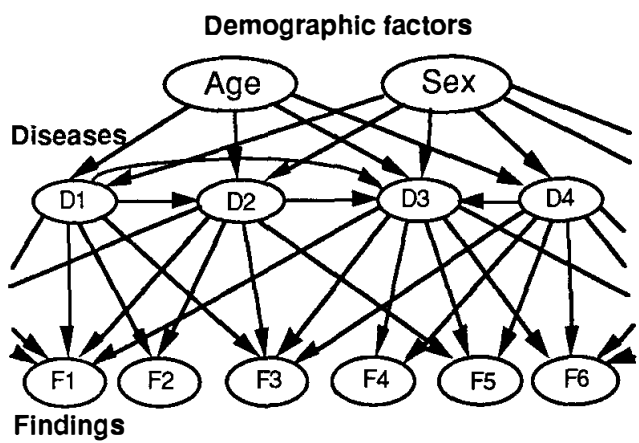

Figure 3: Belief net with causative factors and disease dependencies.

The second stage is to assign probabilities to this framework, either derived from the QMR numbers, or elsewhere. Heckerman & Miller (1986) have demonstrated a fairly reliable monotonic correspondence between the frequency numbers and $P(f/d)$, the link probabilities of a finding $f$ given only disease $d$. Since there are over 40,000 frequencies in QMR, the ability to use a direct mapping does a great deal to ease the reformulation process by avoiding the need to reassess all the disease-finding relationships. We have also developed a mapping from imports to leak probabilities. Finally, our probabilistic representation requires prior probabilities or prevalence rates for each disease, a quantity with no correspondence in the INTERNIST-1/QMR knowledge base. These were estimated from data compiled by the National Center for Health Statistics on the basis of hospital discharges, conditional on the specified demographic (age and sex) categories. In summary, the qualitative independence assumptions of BN2O, together with the link probabilities, leak probabilities, and disease probabilities conditional on age and sex, specify a reformulation of QMR in coherent probabilistic form.

## 4 INFERENCE ALGORITHMS

Given this BN2O representation, is there a tractable method for diagnostic inference? To compute the exact posterior probability of any hypothesis, we need to compute the sum of relative probabilities of all hypotheses. Since the set of complete hypotheses (disease combinations) is the powerset of the set of diseases and has cardinality of $2^{576}$, this may seem a rather daunting prospect. We have explored at least three different approaches for diagnostic inference for this class of networks. These include an exact method (Quickscore), and two approximate methods, one using a forward sampling or simulation scheme, (likelihood weighting), and one using search of the hypothesis tree with probability bounding (TopN).

The QuickScore algorithm (Heckerman, 1989) uses an ingenious rearrangement of the summation. Its complexity is polynomial in the number of diseases but is exponential in the number of findings observed. In practice it can score cases with 12 findings in about 10 minutes (Lightspeed Pascal on a Mac IIci), but it becomes too slow if there are many more findings. Since BN2O has large numbers of intersecting loops, exact methods seem unlikely to be tractable for larger problems.

Likelihood weighting (Shachter & Peot, 1989; Fung & Chang, 1989) is a development of logic sampling (Henrion, 1988) in which each randomly generated hypothesis is weighted by the likelihood of the observed findings conditional on the hypothesis. Further efficiency is achieved by using importance sampling, in which the sampling probabilities of diseases are iteratively adjusted to reflect the evolving estimate of their actual probabilities. The S algorithm (Shwe & Cooper, 1990) initializes the probabilities with a version of tabular Bayes (assuming mutual exclusivity of diseases) as a starting point for sampling. This version converges to reasonable estimates of the posterior probabilities in about 40,000 samples taking an average of 94 minutes for the SAM cases (on a Macintosh IIci).

The TopN algorithm takes a quite different approach, searching among hypotheses, that is complete instantiations of the diseases. It relies for its efficiency on the assumption that of the vast ($2^{576}$ for QMR-BN) set of possible hypotheses, only a tiny fraction of them account for most of the probability mass. Hypotheses with more than a few diseases (five or six at most) have negligible probabilities, since the improbability of that many diseases rapidly outweighs any possible improvement in explaining the observed findings. The second key idea is that, even though computing the *absolute* posterior probability of a hypothesis is intractable in general (requiring summing over all hypotheses), it is easy to compute the *relative* probabilities of two hypotheses (see also Cooper 1984; de Kleer & Williams, 1986; and Peng & Reggia 1987a). The third key element is an admissability heuristic to prune paths that cannot led to the most probable hypothesis (or most probable N hypotheses, hence the name TopN), so that only a small part of the space need be searched. A fourth element are some theorems that allow bounding of the sum of relative probabilities of all extensions of each hypothesis, and hence allow obtaining bounds on the absolute probabilities of hypotheses without examining them all. In the following I will give more detail on these, with some extensions and generalizations of previous results.

---

[2] Reggia and Peng (1987a) term this the causation event and notate it as f:d.

## 5 NOTATION:

I will use the common convention that lower case letters, such as $d$, refer to variables, with uppercase, D and $\overline{D}$, referring to the events $d$=true and $d$=false, respectively. Analogously, if $h$ is a set of diseases, then H denotes the event that all diseases in $h$ are true (present), and $\overline{H}$ denotes the event that all diseases in $h$ are false (absent): [2]

$$H = \bigcup_{\forall d \in h} D, \quad G = \bigcup_{\forall d \in g} \overline{D}$$

Given a set of diseases, $\Delta = \{d_1, d_2, \ldots d_n\}$, a *complete* hypothesis is an event that assigns a value, true or false, to every disease in $\Delta$. A *partial* hypothesis assigns a value to a proper subset of the diseases in $\Delta$, leaving the rest unspecified. If $h \subset \Delta$, then H is a partial hypothesis, since diseases not in $h$ remain unspecified.

Adjacency of events denotes conjunction. So the event HG specifies that all diseases in $h$ are present, all those in $g$ are absent, and the rest unspecified. (We assume $h \cap g = \varnothing$.)

Underlining makes a complete hypothesis from a partial one, assigning absent to all diseases not specified. Thus $\underline{H}$ denotes the event that all diseases in $h$ are present and all others in $\Delta$ absent:

$$\underline{H} = \bigcup_{\forall d \in h} D \cup \bigcup_{\forall d \notin h} \overline{D}$$

---

[2] Note that $\overline{H}$ is not equivalent to $\tilde{H}$, the event that at least one of the diseases in $h$ is absent.

## 6 RELATIVE PROBABILITY AND MARGINAL EXPLANATORY POWER

We define $h_O = \varnothing$ as the empty set of diseases, and $\underline{H}_O$ is the corresponding event that no disease from $\Delta$ is present. The *relative probability* of a hypothesis $\underline{H}$ is the ratio of the posterior probability of $\underline{H}$ given findings F to the posterior probability of hypothesis $\underline{H}_O$:

$$R(\underline{H}) = \frac{P(\underline{H} \mid F)}{P(\underline{H}_O \mid F)} = \frac{P(\underline{H}\ F)}{P(\underline{H}_O\ F)} = \frac{R(\underline{H})}{R(\underline{H}_O)} \qquad [1]$$

TopN starts its search from $h_O$, extending it by adding one disease at a time. To generate the next candidate hypothesis, it adds to the current hypothesis the disease which leads to the largest relative probability. To identify the $n$ most probable hypotheses (hence "TopN"), it applies an admissibility heuristic, which abandons a search path when it provably cannot lead to any hypothesis more probable than the $n$th best so far. TopN's admissibility criterion is based on the concept of Marginal Explanatory Power (MEP).

**Definition (MEP):** The *Marginal Explanatory Power* (MEP) of a disease $d$ with respect to a hypothesis set of diseases, $h$, is the ratio of the posterior probability of the extended hypothesis $h \cup d$ to the posterior of $h$ alone:

$$MEP(D, H) = \frac{P(HD \mid F)}{P(H \mid F)} = \frac{R(HD)}{R(H)} \qquad [2]$$

The MEP is a measure of the increase or decrease (according to whether it is greater or less than 1) in the degree to which the hypothesis explains the findings F due to the addition of $d$. The use of the MEP as the basis of an admissable search heuristic depends on the following result (Henrion 1990a):

**Theorem 1a (declining MEP):** Given a BN2O network, for any disease $d$, and disease sets $h$ and $g$, the marginal explanatory power (MEP) of $d$ with respect to $h$ cannot be less than the MEP of $d$ for any extension $h \cup g$, i.e.

$$MEP(D, H) \geq MEP(D, HG) \qquad [3]$$

When searching for the most probable hypothesis from current hypothesis $h$, if MEP(D, H) $\leq$ 1 then $d$ can be eliminated as a path for exploring as an extension to H, since it cannot lead to a more probable hypothesis. It can also be eliminated as a candidate for extending other extensions of H. Thus the only diseases which need to be considered as extensions of H are those for which MEP(D, H) > 1.

## 7 NEGATIVE PRODUCT SYNERGY AND THE MEP THEOREM

It turns out that Theorem 1a does not require the leaky noisy OR assumption 3 of BN2O, assumed in Henrion

(1990a); a weaker assumption, negative product synergy will suffice. First, we define this property, and then show the more general version of the theorem.

**Definition 2a (two cause NPS):** Suppose there are two propositions, $d$ and $e$, and other variable(s) $x$, that influence finding F according to the conditional probability distribution $P(F| d\ e\ x)$, there is *negative product synergy* in the influence of $d$ and $e$ on $f$, iff

$$\frac{P(F|DE\ x)}{P(F|\bar{D}E\ x)} \leq \frac{P(F|D\bar{E}\ x)}{P(F|\bar{D}\bar{E}\ x)} \quad \forall x. \quad [4]$$

This is the condition required for disease $d$ to "explain away" the evidence F, that is, given F, there is a negative influence between $d$ and $e$ (Henrion & Druzdzel, 1990):

$$P(E|D\ F\ x) \leq P(E|\bar{D}\ F\ x) \quad \forall x. \quad [5]$$

It is simple to show that the noisy OR (with or without leaks) exhibits negative product synergy, and so gives rise to this explaining away phenomenon.

Wellman and Henrion (1991) generalize the definition of product synergy for $n$-ary variables, and discuss its relation to additive synergy. Here we generalize the definition in a different way to apply where there are more than two variables which together influence another variable:

**Definition 2b (n cause NPS):** Consider a set $\Delta$ of propositions which influence finding F, as specified by conditional probability distribution $P(F|\Delta)$. The influence exhibits *negative product synergy*, iff for any sets of propositions $x,y,z \subseteq \Delta$, there is negative product synergy between $x$ and $y$ given $z$, i.e.

$$\frac{P(F|XYZ)}{P(F|YZ)} \leq \frac{P(F|XZ)}{P(F|Z)}. \quad [6]$$

**Assumption 4 (POS):** The influence of every disease $d$ on every finding $f$ is positive, that is, for any set of diseases $h$ not containing $d$,

$$P(F|DH) \geq P(F|\bar{D}H), \forall h \subset \Delta, \text{ where } d \notin h$$

Since the inequality is weak, this also allows diseases and findings to be unlinked (independent). Positive influence from disease to finding is an automatic consequence of Assumption 3, the leaky noisy ORs, but not of negative product synergy.

We can now define a class of bipartite belief nets that generalizes the leaky noisy OR of BN2O to positive links with negative product synergy:

**Definition 3 (BN2NPS):** A bipartite network is said to be BN2NPS if it satisfies Assumption 1 (marginally independent diseases), Assumption 2 (conditionally independent findings), Assumption 4 (positive links), and negative product synergy (NPS) in the influence of the diseases on each finding.

We can now obtain a generalization of Theorem 1a, which applies to BN2NPS:

**Theorem 1b (declining MEP):** Given a BN2NPS network, then, for any disease subsets $x$, $y$, $z$ of $\Delta$, the complete set of diseases, the marginal explanatory power (MEP) of $x$ with respect to $z$ cannot be less than the MEP of $x$ for any extension $y \cup z$, i.e.

$$\text{MEP}(X, Z) \geq \text{MEP}(X, YZ) \quad [7]$$

**Proof:** Taking the ratio of the two sides, and substituting the definition of MEP [2],

$$\frac{\text{MEP}(X, Z)}{\text{MEP}(X, YZ)} = \frac{P(F\ XZ)\ P(F\ YZ)}{P(F\ Z)\ P(F\ XYZ)}$$

$$= \frac{P(F|XZ)\ P(F|YZ)}{P(F|Z)\ P(F|XYZ)} \times \frac{P(XZ)\ P(YZ)}{P(Z)\ P(XYZ)} \quad [8]$$

From the definition of $n$ cause negative product synergy [6] above, we know the first term of the produce above is $\geq 1$. From the marginal independence of diseases, we know that

$$P(XZ) = P(Z) \prod_{d \in z} O(D), \text{ where } O(D) = \frac{P(D)}{1-P(D)}$$

Expanding $P(YZ)$ and $P(XYZ)$ similarly in the second term, the top and bottom cancel out. Hence we are left with the entire ratio as $\geq 1$. QED.

## 8 BOUNDS ON THE PROBABILITY OF EXTENSIONS

We want not just to identify the most probable hypotheses using their relative probabilities, but to obtain bounds on their absolute probabilities. To do this we need to obtain bounds on the relative probabilities of all the extensions of hypotheses in the search tree, so that we can put bound on the contributions of all the hypotheses we do not examine explicitly.

So far we have considered only complete hypotheses, such as $\underline{H}$. The relative probability of a partial hypothesis H is the sum of the relative probabilities of all complete extensions of H, that is all complete hypotheses in which all diseases in $h$ are present, that is,

$$R(H) = \sum_{\forall s \supseteq h} R(\underline{S}) \quad [9]$$

We also need the relative probabilities of partial hypotheses that contain excluded diseases, such as:

$$R(H\bar{G}) = \sum_{\forall s \text{ where } g^c \supseteq s \supseteq h} R(\underline{S}), \quad [10]$$

where $g^c$ is the complement of $g$, i.e. the set of diseases in $\Delta$ but not in $g$.

The following result gives an upper bound for the relative probability of a partial hypothesis $h$ excluding diseases in $g$. It gives it in terms of the relative probability of the corresponding complete hypothesis and the MEP for candidate extension diseases $d$ with respect to $h$, which are relatively easy to compute:

**Theorem 2 (UB1):**

$$R(H\overline{G}) \leq R(\underline{H}) \prod_{\forall d \notin h \cup g}[1+\text{MEP}(D, H)]. \quad [11]$$

This follows from the observation that that at most there is no overlap between the findings explained by each disease, and so the MEP(D, H) for each disease d is the same, no matter how many other diseases are in the hypothesis $h$ it is extending. It is a generalization of Theorem 2 given in Henrion (1990c) for the BN2O assumptions. The complete proof relies on the Declining MEP Theorem 1b, and so it also follows from the more relaxed BN2NPS assumptions.

$R(\underline{H})$ provides a simple lower bound (LB1) for $R(H\overline{G})$. This bound would be attained if all proper extensions $s \supset h$ had probability $R(\underline{S})=0$ (Henrion, 1990c).

An higher lower bound is given by the following:

**Theorem 3 (LB2):**

$$R(H\overline{G}) \geq R(\underline{H}) \prod_{\forall d \notin h \cup g} \frac{1}{1-P(D)}. \quad [12]$$

This follows from Assumption 4 of positive influences, that extending a hypothesis $h$ by disease $d$ cannot reduce the likelihood of evidence F, that is $P(F|\underline{HD}) \geq P(F|\underline{H})$.

There are often diseases $d$ which explain nothing more than hypothesis $h$, that is for which $P(F|\underline{HD})=P(F|\underline{H})$. Since these diseases are independent of the rest conditional on H, it is possible to factor out their contributions to a partial hypothesis H$\overline{G}$ thus:

**Theorem 4 (Factoring independents):**

$$R(H\overline{G}) = R(H\overline{GW}) \prod_{\forall d \in w} \frac{1}{1-P(D)},$$

where $w=\{d : P(F|\underline{HD})=P(F|\underline{H})\}$.

This allows us to remove all such independent (non-explanatory) diseases, $w$, from the candidate list as extensions of $h$, while accounting for their contribution. Note that some diseases have relatively high priors (e.g. peptic ulcer with prior 1.6%) and so are not infrequently among the top ten hypotheses even if there is no specific evidence for them. Application of this result prevents them from cluttering up the search process.

Unfortunately the upper bound UB1 is not always a good guide when there are many diseases each of which can explain a lot relative to $H_0$, i.e. MEP(D, $H_0$)>>1. In the beginning of the search in a case with twenty or more positive findings, UB1 can be very large, for example overflowing an 8 byte floating point number (>$10^{300}$), unless computed as logs. An upper bound avoids this tendency is given by:

**Theorem 5 (UB2):**

$$R(H\overline{G}) \leq R(\underline{H}) + \frac{P(H\overline{G}) - P(\underline{H})}{P(F| H_0)P(H_0)} \quad [13]$$

where $P(H\overline{G}) - P(\underline{H})$

$$= \prod_{d \in h} P(D) \prod_{d \in g}[1-P(D)] - \prod_{d \in h} P(D) \prod_{d \in h^c}[1-P(D)]$$

$$= \prod_{d \in h} P(D) \left[ \prod_{d \in g}[1-P(D)] - \frac{P(H_0)}{\prod_{d \in h}[1-P(D)]} \right].$$

This is based on the observation that at most any extension D to H will completely explain all findings, that is $P(F|\underline{DH}) \leq 1$. This bound is complementary to UB1, with use early in the search in cases with many positive findings.

## 9 SEARCH METHOD

The search uses a best-first approach, where "best" means the candidate partial hypothesis with the greatest possible contribution to uncertainty about the relative posterior probability. This uncertainty is measured as the *maximum error*, the difference between the lower bound 2 and the least of the upper bounds:

$$\text{MaxErr}(h) = \text{Min}(\text{UB1}(h), \text{UB2}(h)) - \text{LB2}(h) \quad [14]$$

We order the candidate hypotheses by MaxErr and select the top one as the next one to expand. This is the one for which expansion has the largest scope for reducing its contribution to the overall uncertainty about the relative probability of all unsearched hypotheses. Each time a hypothesis is expanded, this reduces the bounds on its parents. Search terminates, either when the MaxErr is less than a criterion, **Pmin**, expressed as a fraction of the upper bound on the total relative probability, or when the search runs out of space for the hypothesis tree. As in most best-first or A* searches, the algorithm is liable to be memory bound, running out of space before running out of time.

## 10 OBTAINING ABSOLUTE PROBABILITIES

So far we have obtained bounds on the relative probability of a variety of partial hypotheses, including LBR(H), UBR(H) for each hypothesis H in the search tree, each disease D, LBR(D), UBR(D), and $H_0$. Note that the partial hypothesis $H_0$ is all extensions of the no disease hypothesis, i.e. all possible hypotheses, so $P(H_0) = 1$.

$$R(H_0) = \sum_{\forall s \supseteq h_0} R(\underline{S})$$

$$= \sum_{\forall s \supseteq h_0} \frac{P(\underline{S}\ F)}{P(\underline{H_0}\ F)} = \frac{P(F)}{P(\underline{H_0}\ F)} \quad [15]$$

Hence, $P(F) = R(H_0) P(\underline{H_0}\ F)$ [16]

The posterior probability of any partial hypothesis H is

$$P(H|F) = \frac{P(H\ F)}{P(F)}$$

Substituting in from the definition of relative probability $P(H\ F) = R(H) P(\underline{H_0}\ F)$ and [16] we get

$$P(H|F) = \frac{R(H)}{R(H_0)} . \quad [17]$$

The upper bound for this is when R(H) is at its upper bound UBR(H) and $R(H_0)$ is lower bound $LBR(H_0)$, but note that since the partial hypothesis $H_0$ includes H, we need replace LBR(H) as a component of $LBR(H_0)$ by the the upper bound of H in the denominator too. Thus, we get the upper bound on the posterior probability of H is:

$$UBP(H|F) = \frac{UBR(H)}{LBR(H_0) - LBR(H) + UBR(H)} \quad [18]$$

and similarly the lower bound is

$$LBP(H|F) = \frac{LBR(H)}{UBR(H_0) - UBR(H) + LBR(H)} \quad [19]$$

The maximum total error due to probability of hypotheses not examined in the search is given by

$$\frac{UBR(H_0) - LBR(H_0)}{UBR(H_0)} . \quad [20]$$

TopN also produces a "best" probability estimate for each hypothesis, $h$, defined as the ratio of the sum of the relative probabilities of all complete hypotheses actually examined that contain $h$, to the relative probability of all hypotheses examined, $e$:

$$Best(H) = \frac{\sum_{\forall g \in e\ where\ g \supseteq h} R(\underline{G})}{\sum_{\forall g \in e} R(\underline{G})} \quad [21]$$

This probability estimate is guaranteed to be between the lower and upper bounds on the absolute probability.

## 11 PERFORMANCE OF TOPN:

The QMR-BN research team has assembled cases for testing the performance of alternative inference algorithms. These include 16 cases abstracted from the Scientific American Medicine (SAM) Continuing Medical Education Service. More details of the coding process are given in Shwe et al (1991).

For analysis of timing and accuracy we examined 12 of the 16 SAM cases in which Quickscore can be run for comparison, that is cases with less than 14 positive findings. These cases have an average 9 positive and 11 negative findings. Table 4 gives results for on the performance of TopN for series of runs using a search precision (**Pmin**) of $10^{-5}$. The number of hypotheses examined varies from 277 to 30000. (In two cases search was cut off after 30000 hypotheses due to exhausting memory space.) Since the distributions of hypotheses, time, and precision are highly skewed, Table 1 includes minimum, maximum and median, as well as mean values.

Table 1: Performance on 12 SAM cases using TopN algorithm with a search precision **Pmin** of $10^{-5}$

|  | Min | Max | Mean | Median |
| --- | --- | --- | --- | --- |
| Num of findings | 9 | 28 | 20 | 22 |
| positive | 6 | 14 | 9 | 8 |
| negative | 0 | 20 | 11 | 11 |
| Num of hyps | 277 | 30000 | 11215 | 3794 |
| Run time (secs) | 1.2 | 65.3 | 17.8 | 7.7 |
| Max prob bound | 0.008 | 1.000 | 0.31 | 0.21 |
| St. err. of "best" | <0.00001 | 0.064 | 0.009 | 0.005 |

TopN took an average of 18 seconds (maximum of just over a minute) for the 12 SAM cases. The S sampling algorithm was run for 40,000 samples to achieve adequate convergence for the SAM cases, taking an average of 94 minutes on a Macintosh IIci (about three times faster than the machine used for the Quickscore and TopN runs).

In some cases the maximum probability bound is at or near 1, and quite useless. But it turns out that the actual accuracy of the "best estimate" probabilities is very good when compared with the exact results from QuickScore, with a mean standard error between of 1.2%. Thus it appears that the bounds are highly conservative (much larger than necessary) in most cases. This finding suggests the sampled hypotheses are quite representative in terms of disease probabilities of the unsampled ones. Of course this may not always be true, but it suggests some interesting conjectures about properties of the hypothesis population.

To examine the effect of computational effort on the error, the precision for terminating search **Pmin** was varied by factors of 10 from $10^{-3}$ to $10^{-7}$. Decreasing **Pmin** increases the number of hypotheses explored, and decreases the maximum bound on the probability error. The computation time is approximately linear in the number of hypotheses examined, with 30,000 hypotheses taking about 65 seconds on a plain Macintosh II. Figure 3 shows the effect of increasing the number of hypotheses searched on the error bound for the 16 SAM cases. Most converge satisfactorily according to the error bound by 30,000 hypotheses, but four do not.

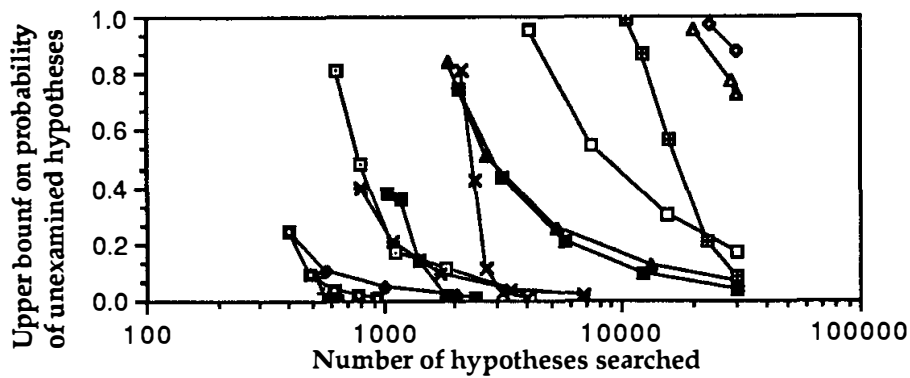

Figure 3: Error bound as maximum probability unaccounted for as a function of the extent of the search for 16 SAM cases.

Figure 3 illustrates how the uncertainty about the computed probabilities decreases as the search of the hypothesis tree is extended, that is as the cumulative probability of all the hypotheses examined approaches one. Thus TopN is an "any time" algorithm: If it is stopped at any point after initialization, it will give bounds on the posterior probabilities; and the longer it runs, the narrower these bounds will be. Given an estimate of the convergence rate, a meta-reasoner could select the run-time to be allocated according to the urgency of the diagnosis, the importance of precision, and the cost of computing.

## CONCLUSIONS

The QMR-BN belief network confronts us with the general intractability of exact algorithms for diagnostic inference. Search-based algorithms such as TopN appear a promising approximate approach for such networks. They may be seen as smarter than forward sampling techniques in that they search specifically to find the most probable instantiations. They rely on exact methods for bounding the error in the resulting probabilities instead of the statistical error estimation methods available for some sampling techniques.

We have presented a variety of results that bound the relative probabilities of partial hypotheses in BN2O and BN2NPS networks. These results illustrate the value of applying methods of qualitative probabilistic analysis, based on knowledge of the signs of influences and synergies. For the QMR-BN project, and no doubt others, there remains a need to develop more general results, for example for networks with prior dependences among diseases. The generality of search-based methods for bounding probabilities remains an open question. It seems unlikely that the kind of bounding results used here will be obtainable for completely general networks, but some further generality may be obtainable from knowledge of qualitative probabilistic properties of other classes of network.

### Acknowledgements

This work was supported in part by the National Science Foundation under grant IRI-8807061 to Carnegie Mellon, and in part by the Rockwell International Science Center.